# How Self-Supervised Learning Can be Used for Fine-Grained Head Pose Estimation?

Mahdi Pourmirzaei[1], Farzaneh Esmaili[1], Ebrahim Mousavi[1], Sasan Karamizadeh[2], Seyedehsamaneh Shojaeilangari[3,*]

{m.poormirzaie, f.esmaili, e.moosavi}@modares.ac.ir, s.karamizadeh@itrc.ac.ir, s.shojaie@irost.ir

Department of Information Technology, Tarbiat Modares University[1]
ICT Research Institute[2]
Biomedical Engineering group, Department of Electrical and Information Technology, Iranian Research Organization for Science and Technology (IROST), P.O. Box 33535111, Tehran, Iran[3]

## Abstract

The cost of head pose labeling is the main challenge of improving the fine-grained Head Pose Estimation (HPE). Although Self-Supervised Learning (SSL) can be a solution to the lack of huge amounts of labeled data, its efficacy for fine-grained HPE is not yet fully explored. This study aims to assess the usage of SSL in fine-grained HPE based on two scenarios: (1) using SSL for weights pre-training procedure, and (2) leveraging auxiliary SSL losses besides HPE. We design a Hybrid Multi-Task Learning (HMTL) architecture based on the ResNet50 backbone in which both strategies are applied. Our experimental results reveal that the combination of both scenarios is the best for HPE. Together, the average error rate is reduced up to 23.1% for AFLW2000 and 14.2% for BIWI benchmark compared to the baseline. Moreover, it is found that some SSL methods are more suitable for transfer learning, while others may be effective when they are considered as auxiliary tasks incorporated into supervised learning. Finally, it is shown that by using the proposed HMTL architecture, the average error is reduced with different types of initial weights: random, ImageNet and SSL pre-trained weights.

## 1. Introduction

Facial image analysis and specially Head Pose Estimation (HPE) is considered as one of the most requesting research topics in computer vision applications such as driver monitoring [1], human-computer interaction [2,3], human behavior analysis [4] and face recognition [5]. Although many techniques have been developed in recent years [6–8] to improve the HPE systems in real world unconstrained scenarios such as occlusion and illumination variation [8], it is still an open research challenge.

In general, there have been two major strategies for HPE [9]. The first one is landmark-based approaches that estimate the head pose using the facial keypoints. Despite great progress in keypoint detection with the advent of deep learning, head pose recovery with this strategy is fundamentally a two steps process with frequent failure possibilities. For instance, if satisfactory keypoints fail to be detected, then pose estimation is impossible [6]. The second strategy is based on end-to-end learning from images in which a neural network tries to find the head pose angles from facial images directly. It is reported that the latter has outperformed landmark-based approaches significantly [6,7,10–12].

Nevertheless, the performance of end-to-end learning is usually constrained by the amount of labeled data. Unlike many other computer vision tasks, collecting the large amounts of labeled data for HPE is challenging due to the complexity and time consuming process of labeling that may adversely impact on quality of annotating. However, recently, with the impressive progress of Self-Supervised Learning (SSL) [13–15], unsupervised learning has gained a lot of attention again to tackle the requiring massive annotated data for end-to-end fashion of HPE. Indeed, SSL is mostly applied as a pre-training step, in which the common end-to-end learning methods are using them to fine-tune the pre-trained weights. However, those methods are more useful for classification tasks, but not suitable for fine-grained problems like HPE. In our opinion, with SSL pre-trainings on head pose images, a model at best can learn race or identity features from images while those representations provide little information about the position of heads. Also, another problem of current SSL techniques is that they need a huge amount of data and heavy computations to work well. It means that for a task like HPE with quite limited resources, those approaches are not supposed to perform successfully. We think that SSL can be helpful for fine-grained tasks in a different setting.

To clarify, in a study [16], it is shown that self-supervised pre-text tasks could extract good features for downstream tasks even if it is applied on one image. In fact, by doing a pre-training, middle layers show valuable features concerning final layers. In other words, a sudden drop in linear feature evaluation can also be observed when moving from low to high level layers. From another view, this issue could be looked through Multi-Task Learning (MTL) settings. Before a certain point, two tasks enhancing shareable features for each other and after that layer, they start hurting each other [17,18]. It is described as cooperation and competition between tasks. So, this article tries to solve a Supervised Learning (SL) task with auxiliary self-supervised tasks simultaneously in the form of MTL.

---

[*] Corresponding author

In this study, we explored the application of SSL for HPE task in two frameworks: (1) using it as the weights pre-training, (2) leveraging auxiliary SSL losses besides of SL ones at the same training procedure named Hybrid Multi-Task Learning (HMTL) [19]. We then compared these two frameworks using puzzling, rotation as the SSL pre-text tasks [20] and Barlow Twins as the non-contrastive SSL [21]. Thereafter, we examined the impact of adding auxiliary SSL losses to the fine-grained head pose ones in different settings. It is found that our proposed HMTL architecture performed differently compared to utilizing SSL in the conventional protocol i.e., pre-training and then fine-tuning. In simple words, if you have a supervised task A and a self-supervised task B, there is a difference between results of using B as pre-training, with using B as an auxiliary loss besides A.

All together, we designed a multi-task neural network based on ResNet50 [22] architecture to couple with the self-supervised auxiliary losses. As a matter of fact, we investigated various representation of SSL auxiliary losses to reduce the pose estimation error rate. In addition, we compared the impact of initial weights on the proposed multi-task architecture "HMTL" and showed that the error was reduced with all kinds of initializer such as random weights, ImageNet weights and SSL pre-trained weights. Our proposed method was trained on 300W-LP [23] and evaluated on AFLW2000 [24], BIWI [25] and ETH-XGaze [26] databases.

The contributions of this paper are summarized as follows:

1. We defined and compared two different frameworks of using self-supervised learning for HPE.
2. We investigated the impact of merging self-supervised learning losses with HPE ones by designing a unified architecture.
3. Using our proposed architecture, the effect of self-supervised pre-training and supervised ImageNet pre-training has been scrutinized.
4. The difference of our proposed hybrid learning with supervised learning was explored in a low data regime.

## 1.2 Related works

Among several deep learning models, convolutional neural networks (CNNs) have achieved promising results in pose estimation [27]. Although, HPE is a regression problem, since the advent of HopeNet [6], recent studies have combined the regression with the classification task [11,28]. HopeNet utilizes multi-task loss function for all three Euler angles (yaw, pitch, and roll) separately. Its loss function consists of two parts; the first one is a pose bin classification and the second one is the regression part calculated by expectation of pose classification (Fig 1).

In the research [8], the authors reformulated the HPE problem as a label distribution learning task, considering each facial image as an instance affiliated with a Gaussian label distribution rather than a single label, then they built a CNN network with muti-loss function.

Furthermore, a work [10] developed a HPE model using two stage ensembles with top-k regression. The first stage was a binned classification and the second stage, unlike HopeNet, used the average of top-k classes to find regression values instead of expectation of all bins. They showed that the efficiency of bin classification was dependent on angle distance when using CNNs at the first stage.

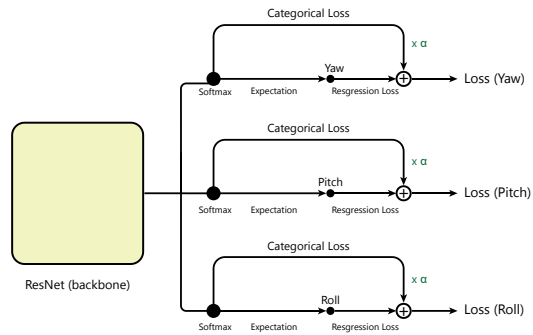

Fig 1 Original HopeNet architecture with three yaw, pitch and roll heads. Each head has a separate loss function which utilizes both categorical and regresion into account. In fact, HopeNet architecture consists of a backbone and a customized loss function.

In another study [11], the WHE-Net was proposed as a network to predict the full-range of head yaws, which performed well on frontal views. Afterwards, the authors created a wrapped loss function to enhance the performance of estimation for anterior views in full-range HPE which properly handled wrapping of yaw. The proposed wrapped loss function, instead of penalizing angle straight, penalized the minimal rotation angle that was necessary.

A work [9] tried to enhance the HPE by designing a method with two stages. First stage contained a neural network with one regression head and four regression classification heads. With using offsets of bounding box, an ensemble of offsets was built. Second stage performed knowledge distillation from the ensemble of offsets of the base neural network.

Although a large number of unlabeled images under various poses is freely available on the Internet, none of the above-mentioned studies address the HPE's primary challenge, which is to use techniques that do not incorporate the pose labels to improve performance.

## 2. Self-supervised learning

SSL approaches aim to learn inherent features from unlabeled data and can be categorized into three groups [29]:

1. Pretext task learning
2. Contrastive learning
3. Non-contrastive learning

The early techniques of SSL are based on auxiliary pretext tasks in such a way to learn the semantic representation via handcrafted pretext tasks (i.e. rotation [30], colorization [31] and jigsaw puzzling [20]) over unlabeled data and then apply the learned representation for downstream tasks. However, the most recent techniques focus on contrastive [20,32] and non-constructive [13,14,21] learning by using positive or negative image pairs to learn the similarities between different views of the same image and vice versa.

The SSL methods have recently achieved great success in the literature [13,32]; however, they have some drawbacks that make them unsuitable for real-world applications. For example, the current techniques require large amounts of computation resources and data to operate well [33] as well as large batch sizes to train [32]. In addition, they tend to perform worse on fine-grained visual classification tasks in comparison with supervised learning methods [33,34].

To the best of our knowledge, there are few studies related to SSL for head view point estimation [19,35]. Pourmirzaei et al. [19] tried to incorporate the SSL auxiliary losses into SL in a MTL manner. They merged Self-Supervised Heads (SSHs) of pre-text tasks with fine-grained facial emotion recognition and HPE. Nonetheless, their study on HPE was limited. Besides limited research underlying SSL-based HPE, there are some related works into similar fields like body pose [36,37] and hand pose estimation [38]. Moreover, Kundu et al. [37], proposed an analysis-by-synthesis framework for viewpoint learning in a SSL setting utilizing cycle-consistency losses between a viewpoint estimation and a viewpoint aware synthesis network.

There are evidences [39–41] showing SSL is not limited to pre-training stage only, it can be used in a co-training setting for supervised tasks which can reduce the limitation of SSL pre-training in visual tasks. Yun et al. [39] developed a semi-supervised learning framework that benefited the co-training idea using self-supervised techniques for visual representations. This method utilized exemplar and rotation self-supervision techniques dealing with limited annotated data, and their model performance was improved by using only 10% labeled data and 90% unlabeled data. Dahiya at al. [40] found that using cross-modal self-supervision for pre-training is a good idea and could improve the multi-sensory model's performance. They implemented a form of co-training in which the two modalities were correlated but had different types of information about the video. Zhai et al. [41] implemented a self-supervised co-training scheme called Co-training Contrastive Learning of visual Representation (CoCLR) that enhanced a popular loss function named instance-based Info Noise Contrastive Estimation [42] (infoNCE) . Furthermore, they showed that hard positives were being ignored in the self-supervised training, and if these hard positives had been used, then the quality of learned representation would be enhanced enormously.

# 3. Methodology

The framework of designing an architecture for SSL pre-text tasks is described in this section. Note that in this work, HMTL and SL with auxiliary SSL losses refer to the same thing and may be used interchangeably.

## 3.1 Self-supervised tasks

For head pose estimation, we consider three SSL pre-text tasks:

**Puzzling.** Like a related study [19], we apply some sort of puzzling approach by partitioning an image into multiple tiles and shuffling them. A neural network model is then tasked to un-shuffle the image tiles back into the original form. For each region or tile, a classification head is created. For example, 2×2 puzzling needs four SSL heads of classification (Fig 2): $\{H \; \epsilon \; (Region \; 1, \dots, Region \; 4)\}$.

**Rotation.** Images are rotated by the conventional rotation method [30] at n×90 degrees, where n is chosen randomly from $\{0, \dots, 3\}$. It means that an image can be rotated in four directions which are considered as the output label: $\{L \; \epsilon \; (1, \dots, 4)\}$. So, this is a 4-class classification task. However, in contrast to the conventional approach, our rotation method is performed on each tile of an image (Fig 2). This means each piece of an image can rotate independently from another. So, the main rotation approach converts to four (4-class) classification tasks.

**Puzzling-Rotation.** The puzzling-rotation approach is based on the combination of both tasks, where the rotation task is performed on each image tile individually. Therefore, for each region, two tasks are defined; for 2×2 puzzling-rotation, eight self-supervised tasks can be designed. Nevertheless, if we rotate all puzzle pieces, the information regarding head pose will be lost as a result of perturbation. In other words, there can be more than one correct solution for a single sample and this is not good to route head pose information through the network. To overcome this issue, we have to rotate at most two puzzle pieces randomly and the rest must be left unchanged. It is obvious that unchanged pieces take "0" label for rotation task.

## 3.2 Multi-task architecture

Two architectures are proposed in this section: SL, HMTL (SL with auxiliary SSL losses). The first one is an exact replica of the HopeNet [6], whereas the second is an extension of the original HopeNet based on SSL branches.

In order to find the best architecture of HMTL, four flag points in ResNet50 [22] are defined to find the best branching point on the main supervised branch for each self-supervised task. We choose ResNet 50 due to the fact that it has used as the backbone of original HopeNet architecture. As ResNet50 has four main blocks and each block includes sub-blocks of convolution layers, flag points are defined after main blocks (Fig 2). After finding the best dividing point, we will be able to create the best SL + SSL architecture based on the ResNet50 architecture.

$$\mathcal{L}_{Total} = \mathcal{L}_{SL} + \mathcal{L}_{SSL} =$$

$$(\mathcal{L}_{Yaw} + \mathcal{L}_{Pitch}) + \left(\sum_j \mathcal{L}_{Puzzle_j} + \sum_j \mathcal{L}_{Rotation_j}\right) =$$

$$\left(\left(\sqrt{\frac{1}{n}\sum_{j=1}^{n}(\hat{y}-y)^2}\right)_{yaw} + \left(\sqrt{\frac{1}{n}\sum_{j=1}^{n}(\hat{y}-y)^2}\right)_{pitch}\right)$$

$$+ \sum_i^n \left(\left[\sum_{part\;j} y_{i,part\;j} \log(\hat{y}_{i,part\;j})\right]_{puzzle} + \left[\sum_{part\;j} y_{i,part\;j} \log(\hat{y}_{i,part\;j})\right]_{rotation}\right), (1)$$

Where:

- $L_{SL}$: two RMSE losses for SHs
- $L_{SSL}$: categorical cross entropy for puzzling SSHs and rotation SSHs
- $y$: true label
- $\hat{y}$: predicted label
- $n$: number of samples
- $part$: local regions of each image

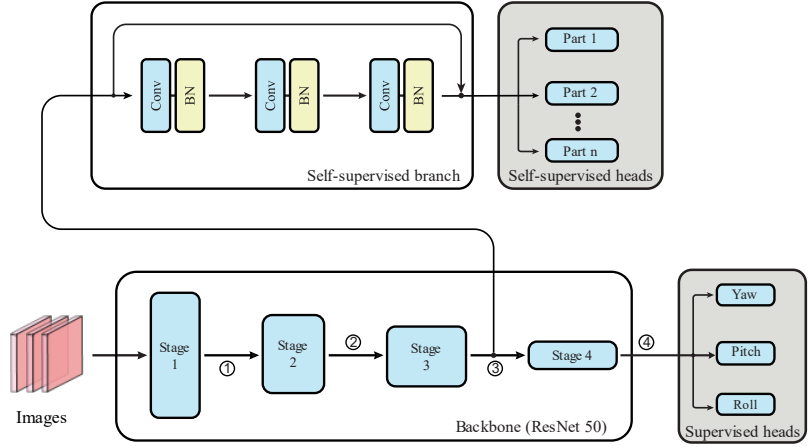

Fig 2 Testing different branches for SHs and SSHs. Each number is a point for self-supervised branch separation.

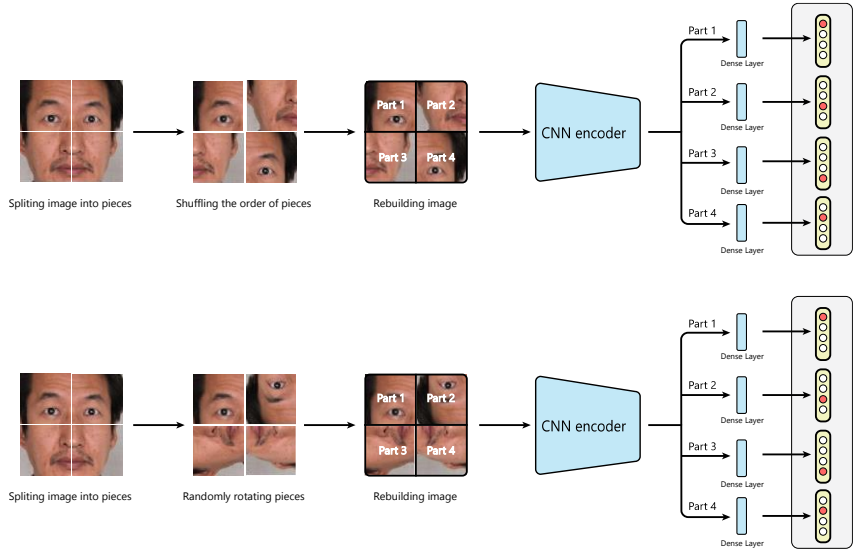

Fig 3 Procedure of SSL puzzling (upper) and rotation (lower) methods. Each output head describes a local position in the image. The puzzling-rotation method is the combination of both. It means for each local region (part), we have two classification heads, one output for the rotation label and the other one for the puzzling label.

Equation 1 shows the main loss function of head pose estimation with auxiliary losses of self-supervised puzzling-rotation method in order to find the best architecture. In this section, we want to evaluate the impact of auxiliary losses on the simplest and commonly used regression head pose losses which are linear layers on top of the backbone with RMSE loss functions. Because we do not want to bias on a specific method such as HopeNet when selecting the best multi-task architecture.

## 3.3 Head pose estimation

In this section, we look through the differences among three approaches on supervised fine-grained head pose estimation:

**SSL pre-training weights.** In this situation, at first, the backbone which is a ResNet 50 model is pre-trained using self-supervised pre-text tasks and then should be fine-tuned on the head pose labels.

**ImageNet pre-training weights.** This is very similar to the previous step, but we use ImageNet weights for the backbone. In other words, we consider transfer learning based on the pre-trained ImageNet weights as the initial weights of backbone.

**Training with auxiliary SSL losses (HMTL).** This is our proposed approach that uses self-supervised auxiliary losses to help supervised loss functions. The architecture is based on the HopeNet architecture with additional layers based on the best architecture we aim to find in Section 3.1. Since, the backbone of this approach is based on ResNet50, besides random weights initialization, we would be able to use ImageNet and SSL pre-training weights for it. Since the self-supervised auxiliary tasks are classification problems, in this approach, all losses are considered categorical cross entropy.

### 3.4 Datasets

In this paper, datasets below have been used:

**300W-LP** [23]**.** It consists the extended version of 300W dataset, in which the images of several different databases of faces along with facial points are gathered together. 300W-LP contains 61225 images.

**AFLW2000** [24]**.** It consists of 2000 images from the AFLW database, which are accurately labeled by a 3D modeling approach. This database contains facial landmarks, as well as the labels of three head pose Euler angles in a precise way, for this reason, it is considered as the test set in this research.

**BIWI** [25]**.** It was collected in a laboratory environment by recording RGB-D videos of different human subjects in different poses using Kinect v2. These videos contain about 15,000 frames created at different head angles. This dataset is usually used as a benchmark for head angle estimation using depth estimation methods. Obviously, in this research, only color images are used to estimate the head angle.

**ETH-XGaze** [26]**.** This database consists of more than one million high resolution images of wide head angles from 110 subjects. Images and labels of participants were recorded with customized and accurate hardware settings. This database was primarily created for gaze estimation. But in addition to that, two head pose angles (yaw and pitch) of the participants have been calculated and made available in ETH-XGaze.

## 4. Experiments

As we described in Section 3.2, the ResNet50 architecture is used as the encoder. The input image size for all designed networks is set to 224×224×3. For training all models, an Adabelief optimizer [43] with batch size of 64 is utilized. Also, mixed precision is applied to speed up the training process.

Note that the experiments were implemented by TensorFlow 2 framework in python and the networks were trained using 2×RTX 2070 8GB GPU cards.

### 4.1 Exploring the best HMTL architecture for head pose estimation

Here, at first, the original ResNet50 architecture is considered as the backbone and then, four flags are defined on it as described in Section 3.2. ETH-XGaze dataset is selected for this experiment. Note that this data is sorted by the number of human subjects and consists of yaw and pitch label for head pose estimation and does not include the label of roll. Due to the huge amounts of images and numerous experiments, one subject from the training set (including 10,000 samples) is selected to minimize the time and computational cost of experiments. In addition, we would be able to measure the performance of different architectures more clearly in the low data regime.

Since, ETH-XGaze has no publicly available validation or test sets, we have randomly selected and separated three subjects from the training subjects (id of 108, 109 and 111) as the validation data which includes 30,000 images. Moreover, we have applied a series of data augmentation like random zooming, random hue transformations and dropout [44] to prevent overfitting. On the SHs and SSHs, dropout layers with values of 0.4 and 0.2 are used respectively. Each SSL pre-text task branch consists of a residual convolution block in which three ConvNet layers with batch normalization are placed inside a skip connection. After the convolution block, for each local zone of the input images, a dense layer followed by a softmax function is placed as a SSH with respect to each pre-text task (rotation and puzzle). Both the rotation and puzzle pre-text branches are the same. We aim to find the best architecture of HMTL using three SSL pre-text tasks:

- 2×2 Puzzle
- 2×2 Rotation
- 2×2 Puzzle-Rotation

Need to point that in the Puzzle-Rotation task, each rotation and puzzling branches are designed by different networks i.e., unique weights, however, the location separated from the backbone is the same. Finally, a batch of training samples is represented as $\{(X,Y) = (X, (R_i, P_j); i \in (1, ..., 4) \& j \in (1, ..., 4))\}$ for the Puzzle-Rotation task, $\{(X,Y) = (X, (P_i); i \in (1, ..., 4))\}$ for the Puzzle task and $\{(X,Y) = (X, (R_i); i \in (1, ..., 4))\}$ for the Rotation task. Results are reported in Table 1 and Fig 4.

We have designed two supervised heads belonging to yaw and pitch labels. Each one has been created by a regression dense layer on top of backbone (ResNet) output. All methods have been trained for 110 epochs. In the beginning, learning rate is set to 0.001, and in epoch 30 and 40, it is multiplied by 0.1.

As shown in Fig4, the point number 3 has the minimum average error rate compared to the rest. Based on the results tabulated in Table 1, we can see the yaw error rate drops remarkably when puzzling losses are added to the network. However, the rotation auxiliary task does not help SHs to reduce the error rate, neither alone or combining with puzzling. Consequently, we have selected the output of stage 3 as the best point for adding SSL heads to the backbone and considered it as the proposed HMTL architecture in this paper.

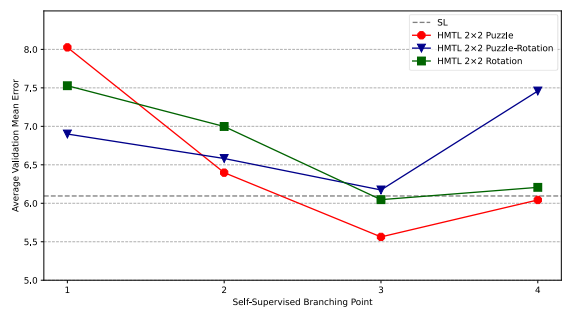

Fig 4 The average of yaw and pitch mean error via different flags with 2×2 puzzling, rotation and puzzling-rotation on the ETH-XGaze images. It shows that that HMTL 2×2 puzzle performed better than others at point 3.

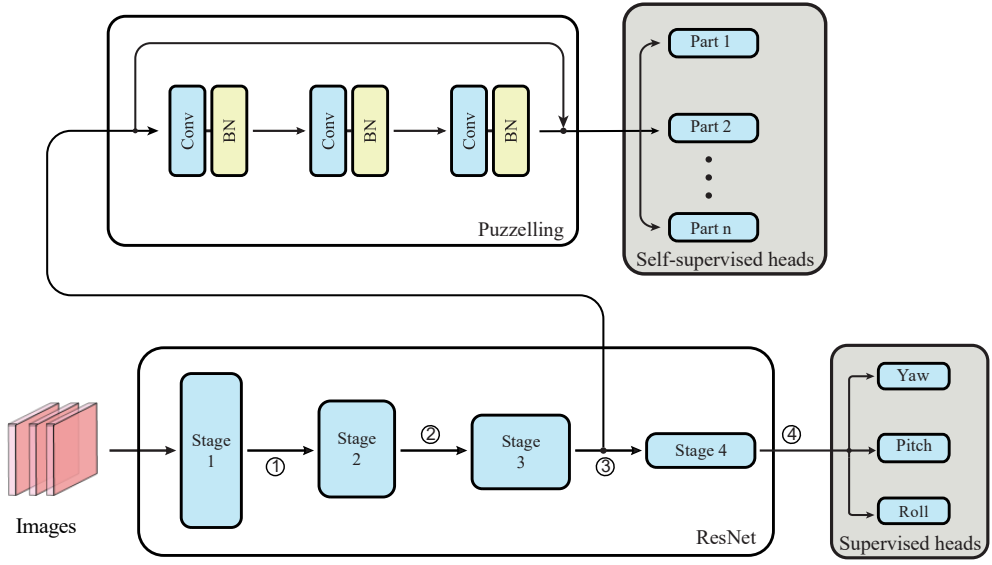

Fig 5 The proposed architecture for adding self-supervised auxiliary losses based on the N×N puzzling pre-text task.

Table 1 Results of mean average errors for Yaw and Pitch via different architectures and SSL losses on the ETH-XGaze images.

| Method | Flag | Yaw (MAE) | Pitch (MAE) | Average |
|---|---|---|---|---|
| HMTL puzzling-rotation 2×2 | 1 | 5.075 | 8.725 | 6.9 |
| HMTL puzzling-rotation 2×2 | 2 | 5.243 | 7.921 | 6.582 |
| HMTL puzzling-rotation 2×2 | 3 | 5.243 | 7.921 | 6.582 |
| HMTL puzzling-rotation 2×2 | 4 | 5.243 | 7.921 | 6.582 |
| HMTL rotation 2×2 | 1 | 4.985 | 10.072 | 7.528 |
| HMTL rotation 2×2 | 2 | 4.373 | 9.623 | 6.998 |
| HMTL rotation 2×2 | 3 | 4.757 | 7.337 | 6.047 |
| HMTL rotation 2×2 | 4 | 4.701 | 7.712 | 6.207 |
| HMTL puzzling 2×2 | 1 | 6.478 | 9.575 | 8.026 |
| HMTL puzzling 2×2 | 2 | 5.667 | 7.13 | 6.398 |
| **HMTL puzzling 2×2** | **3** | **4.272** | **6.852** | **5.563** |
| HMTL puzzling 2×2 | 4 | 4.347 | 7.738 | 6.043 |
| SL | - | 5.338 | 6.852 | 6.095 |

## 4.2 Self-supervised learning for fine-grained head pose estimation

In the previous section, we defined several dividing points for appending the SSL branches to the backbone to find the best point. In this section, we consider the best architecture (flag point 3) as our proposed HMTL architecture for head pose estimation

As mentioned in Section 3.3, our goal was to analyze the impact of three approaches on the original supervised HopeNet architecture:

1. Fine-tuning on ImageNet pre-trained weights
2. Fine-tuning on SSL pre-trained weights
3. Adding SSL auxiliary losses

We need to clarify that the proposed HMTL is the combination of HopeNet and the SSL branches. In other words, it consists of three parts: Backbone (ResNet) + HopeNet loss + SSL branches. As a matter of fact, using pre-training weights on the HMTL network, refer to do pre-training on the backbone (ResNet) alone and subsequently, the SSL branches' weights are initialized randomly.

$$\mathcal{L}_{Total} = \mathcal{L}_{SL} + \mathcal{L}_{SSL} = \mathcal{L}_{Cat-Reg} + \mathcal{L}_{puzzle}$$

$$= (\mathcal{L}_{Cat} + \alpha \mathcal{L}_{Reg}) + \mathcal{L}_{puzzle}$$

$$= \left(-\sum_i y_{Cat_i} \log(\hat{y}_{Cat_i}) + \alpha \sqrt{\frac{1}{n}\sum_{j=1}^{n}(y_{Reg_j} - E(\hat{y}_{Cat})_j)^2}\right)_{yaw}$$

$$+ \left(-\sum_i y_{Cat_i} \log(\hat{y}_{Cat_i}) + \alpha \sqrt{\frac{1}{n}\sum_{j=1}^{n}(y_{Reg_j} - E(\hat{y}_{Cat})_j)^2}\right)_{pitch}$$

$$+ \left(-\sum_i y_{Cat_i} \log(\hat{y}_{Cat_i}) + \alpha \sqrt{\frac{1}{n}\sum_{j=1}^{n}(y_{Reg_j} - E(\hat{y}_{Cat})_j)^2}\right)_{roll}$$

$$+ \sum_i^n \left(\sum_{part\,j} y_{i,part\,j} \log(\hat{y}_{i,part\,j})\right)_{puzzle}, (2)$$

Where:

- $L_{Cat}$: categorical cross entropy for categorical loss
- $L_{Reg}$: RMSE loss functions for yaw, pitch and roll heads
- $\hat{y}_{Cat}$: output of softmax layer
- $y_{Cat}$: true label for categorical loss
- $\alpha$: weight for the regression losses
- $E$: expectation of softmax layer after categorical head output to calculate regression value
- $y$: true label
- $\hat{y}$: predicted label
- $n$: number of samples
- $part$: local regions of each image

Table 2 Results of different methods on the AFLW2000 benchmark. Our SL methods are based on the HopeNet architecture that we implemented and trained. *: indicates that the AFLW2000 benchmark without removing the 31 samples is reported. ┼: indicates that bin size for categorical intervals set to 1. The bin size for the rest is 3. ∩: down sampling transformation to 15X.

| Method | Augmentation | Pre-train weights | Yaw | Pitch | Roll | Average |
|---|---|---|---|---|---|---|
| FAN (12 points) [45] | Unknown | - | 18.273 | 12.604 | 8.998 | 13.292 |
| 3DDFA [23] | Unknown | - | 5.4 | 8.53 | 8.250 | 7.393 |
| RetinaFace R-50 (5 points) [46] | Unknown | - | 5.101 | 9.642 | 3.924 | 6.222 |
| HopeNet [6] | ≈2 | - | 6.47 | 6.559 | 5.436 | 6.155 |
| Hybrid Coarse-Fine [47] | Unknown | - | 4.82 | 6.227 | 5.137 | 5.395 |
| HPE-40 [10] | Unknown | - | 4.87 | 6.18 | 4.8 | 5.28 |
| FSA-Caps-Fusion [12] | ≈2 | - | 4.5 | 6.08 | 4.64 | 5.07 |
| WHENet [11] | ∩ | - | 4.44 | 5.75 | 4.31 | 4.83 |
| TriNet [48] | Unknown | - | 4.198 | 5.767 | 4.042 | 4.669 |
| QuatNet [28] | Unknown | - | 3.97 | 5.61 | 3.92 | **4.5** |
| Img2pose [49] | Unknown | - | 3.426 | 5.034 | 3.278 | **3.913** |
| SL | 1 | - | 5.736 | 5.907 | 4.89 | 5.511 |
| SL | 1 | ETH-XGaze SSL 2×2 rotation | 5.86 | 5.541 | 4.113 | 5.171 |
| SL | 1 | ImageNet | 5.973 | 5.488 | 4.191 | 5.217 |
| SL* | 2 | - | 5.278 | 7.987 | 6.638 | 6.634 |
| SL | 2 | - | 6.221 | 5.569 | 3.984 | 5.258 |
| SL | 2 | ETH-XGaze SSL 2×2 rotation | 5.355 | 5.432 | 4.175 | 4.987 |
| HMTL 2×2 puzzling | 1 | - | 4.22 | 6.065 | 5.007 | 5.094 |
| HMTL 3×3 puzzling | 1 | - | 4.175 | 5.8 | 4.951 | 4.975 |
| HMTL 3×3 puzzling | 1 | ETH-XGaze SSL 2×2 rotation | 3.855 | 6.065 | 4.377 | 4.766 |
| HMTL 3×3 puzzling | 1 | ImageNet | 4.235 | 6.2 | 4.231 | 4.889 |
| HMTL 3×3 puzzling* | 2 | - | 3.742 | 7.796 | 6.496 | 6.011 |
| HMTL 3×3 puzzling | 2 | - | 3.874 | 5.929 | 4.416 | 4.74 |
| HMTL 3×3 puzzling┼ | 2 | - | 3.967 | 5.942 | 4.191 | 4.7 |
| HMTL 3×3 puzzling | 2 | ETH-XGaze SSL 2×2 rotation | 3.682 | 5.919 | 4.316 | **4.639** |
| HMTL 2×2 rotation | 1 | - | 5.998 | 5.604 | 4.31 | 5.304 |
| HMTL 2×2 rotation | 1 | ETH-XGaze SSL 2×2 puzzling | 5.77 | 5.657 | 4.104 | 5.177 |
| HMTL 2×2 puzzling-rotation | 1 | - | 5.554 | 6.211 | 5.312 | 5.692 |

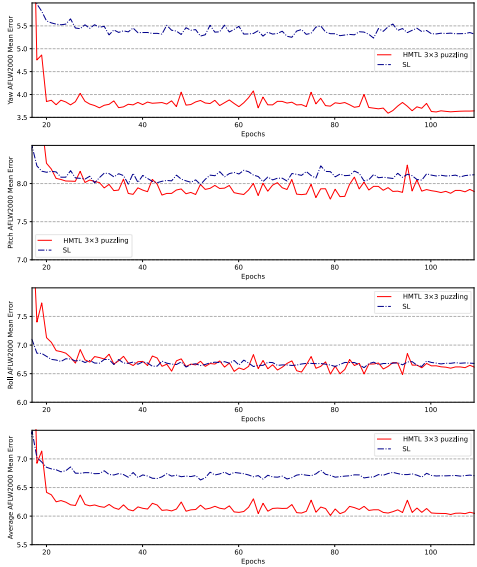

Fig 6 Comparing yaw, pitch and roll validation errors for SL and HMTL on the AFLW2000 images. Here, the SL is identical to the original HopeNet architecture. The augmentation level is set to level 2. Note that for evaluation, AFLW2000 is used w/o removing any samples from it.

Table 3 Results on the BIWI benchmark. The SL methods are based on the HopeNet that we implemented. All of our methods use level two of augmentation and the bin size for the categorical intervals is set to 3.

| Method | Pre-train weights | Yaw | Pitch | Roll | Avg |
|---|---|---|---|---|---|
| HopeNet [6] | - | 4.810 | 6.606 | 3.269 | 4.895 |
| QuatNet [28] | - | 4.01 | 5.49 | 2.93 | 4.14 |
| FSA-Caps-Fusion [12] | - | 4.27 | 4.96 | 2.76 | 4.0 |
| WHENet [11] | - | 3.60 | 4.10 | 2.73 | **3.48** |
| SL | - | 4.322 | 5.94 | 3.113 | 4.458 |
| SL | ImageNet | 4.379 | 5.636 | 3.002 | 4.339 |
| SL | ETH-XGaze SSL 2×2 rotation | 4.298 | 5.891 | 2.782 | 4.323 |
| HMTL 2×2 rotation | - | 4.356 | 5.801 | 3.27 | 4.476 |
| HMTL 3×3 puzzling | - | 4.123 | 5.317 | 3.059 | 4.166 |
| HMTL 3×3 puzzling | ImageNet | 3.988 | 4.952 | 3.01 | 3.983 |
| HMTL 3×3 puzzling | ETH-XGaze SSL 2×2 rotation | 3.864 | 4.641 | 2.962 | **3.822** |

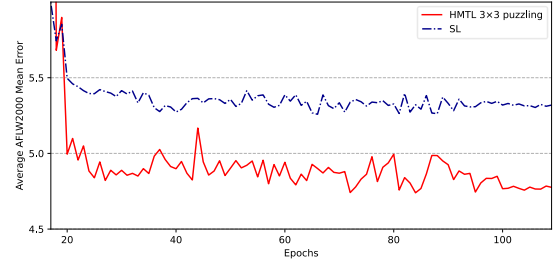

Fig 7 Comparing average MAE for SL and HMTL on the reduced version of AFLW2000 at different epochs. The SL method is based on the HopeNet method. Augmentation level is set to level 2.

In this section, We use 300W-LP images [23] to train our models and both AFLW2000 and BIWI benchmarks to evaluate them. We have trained all models for 110 epochs. All learning rates start with 0.001 and at epoch 20 and 100, they are multiplied by 0.1. Also, two levels of augmentation have been applied. The first level consists of random central zoom and random contrast, and the second level includes all transformations in level one, as well as blurring, down scale resolution and cutout. Moreover, for all supervised and self-supervised heads, dropout layers with values of 0.5 and 0.2 have been used respectively. Since the value of supervised losses are much bigger than each self-supervised losses, all loss function of self-supervised heads, puzzling and rotation, have been multiplied by 50.

The total loss value of the proposed puzzling-based HMTL is formulated by Equation 2. In the HopeNet loss function considered for the SHs in Equation 2, alpha coefficient defines the balance between regression and categorical errors. Similar to the HopeNet's paper, we set it 2.

As mentioned, in this section, SSL pre-training has performed only on the backbone weights (ResNet50). At first, in contrast to baseline method (HopeNet), in AFLW2000 benchmark, we did not remove the samples with the absolute values larger than 99 degrees which includes 31 samples. We observed that those samples had a remarkable impact on the results (Fig 6). Therefore, we consider the reduced version of AFLW2000 benchmark (without those 31 samples) to compare it with other methods fairly (Fig 7). Furthermore, we set the bin size for categorical tasks to both 3 and 1 degrees. Results are shown in Table 2 and 3. Table 2 reveals that auxiliary SSL tasks can significantly reduce the average error rate. Nevertheless, in contrast to the puzzling pre-text task, the rotation auxiliary task does not help SHs.

## 4.3 Barlow twins pre-training

For a comparison, we use Barlow Twins [21] (BT) based SSL to pre-train the backbone of HopeNet. As far as we are concerned, BT is one the best SSL methods which is easy to implement compared to other methods such as DINO [14], BYOL [13], SimCLR [32].

In order to train the ResNet50 based on BT technique, we choose the training set of ETH-XGaze dataset which consists of approximately 0.75 million images. Since BT does not need any negative pairs to train, each image is transformed twice to create two distorted views. To be precise, an image initially is rotated randomly in {0, 90, 180, 270} degrees and then two randomly selected transformations are performed on them. The transformations include random cropping, color jittering, converting to grayscale, random Gaussian noise, Gaussian blurring, cutout and random resizing. Random cropping and random Gaussian noise have been always applied, though color jittering and converting to grayscale are randomly applied with 0.8 and 0.3 probability. Then, one of the blurring and resizing is selected randomly and each one can be applied with 0.2 probability. Finally, cutout is performed on each image separately. Except above transformations, we apply BT pre-training one more time in which random 3×3 puzzling transformation is also included to list of transformations. In order to prohibit the model from shortcut learning, random 3×3 puzzling transformation has been applied before performing cutout.

Table 4 Results of BT and HMTL on the reduced version of AFLW2000. Encoder pre-training is performed on the ETH-XGaze images. All the trainings are done with level 2 augmentation under same hyperparameter settings. HMTL refers to 3×3 puzzling-based HMTL. The LE and FT refer to linear evaluation and fine-tuning respectively. *: indicates that pre-training is done by adding random 3×3 puzzling transformation.

| Method | Encoder pre-training | Yaw (MAE) | Pitch (MAE) | Roll (MAE) | Avg |
|---|---|---|---|---|---|
| HopeNet [6] | - | 6.221 | 5.569 | 3.984 | 5.258 |
| HopeNet [6] (LE) | 2×2 Rotation | 17.198 | 11.694 | 11.775 | 13.556 |
| HopeNet [6] (LE) | 2×2 Puzzling | 17.408 | 11.453 | 11.465 | 13.442 |
| HopeNet [6] (LE) | 3×3 Puzzling | 14.509 | 11.132 | 10.657 | 12.099 |
| HopeNet [6] (LE) | BT [21] | 9.1 | 11.753 | 10.820 | 10.558 |
| HopeNet [6] (LE) | BT* [21] | 10.136 | 9.477 | 9.873 | 9.799 |
| HopeNet [6] (FT) | BT [21] | 4.790 | 5.34 | 3.677 | 4.603 |
| HopeNet [6] (FT) | BT* [21] | 4.56 | 5.289 | 3.596 | 4.481 |
| HMTL | - | 3.874 | 5.929 | 4.416 | 4.74 |
| HMTL (FT) | BT [21] | 3.406 | 5.718 | 4.103 | 4.409 |
| HMTL (FT) | BT* [21] | 3.351 | 5.633 | 3.829 | **4.271** |

Similar to original BT's [21] implementation, here, the encoder (backbone) consists of a ResNet-50 network followed by a projector head. The projector network contains three linear layers, each with 2048 output units. The first two layers of the projector are followed by a batch normalization layer and rectified linear units. The encoder has been trained with batch size of 64 for 64 epochs and the learning rate was 0.001 with cosine decay learning rate scheduling. Next, the backbone of HopeNet and puzzling-based HMTL architecture have been initialized with BT weights. Like previous section, we use the AFLW2000 benchmark for evaluation and 300W-LP images for fine-tuning. All fine-tuning start with 5 epochs of warmup. The results of all experiments are shown in the table 4. Based on the results, HMTL with BT weights has lower average MAE compared to the vanilla fine-tuning on the HopeNet. Table 4 specifies two things: (1) the average error rate in the HMTL approach can improve using other SSL pre-training techniques and (2) it indicates that there is a difference between using SSL as auxiliary losses (HMTL) and using SSL in the conventional form i.e., pre-training weights and then fine-tuning on them.

# 5. Ablation study
## 5.1 Impact of self-supervised losses on supervised heads

Yang et al. [12] showed that local features are important for pose estimation. Their proposed method handles the input as a bag of features from an image and ignores the spatial relationship in the feature map. In other words, it is declared that local spatial features are quite useful to estimate fine-grained pose estimation. Regarding the research done by Yang et al., we encountered a question: how much local and global spatial features are important for HPE?

In this part, we try to find out the answer via doing an ablation test. In fact, we investigated the impact of SSHs on the main supervised task. At first, the original HopeNet model (SL) has been trained by giving puzzled images similar to the input of puzzling-based HMTL. We call it HMTL w/o SSHs (w/o stands for with or without). For instance, HTML 2×2 puzzling w/o SSHs means that the HopeNet architecture has been trained on the 2×2 puzzled images. Then, we have compared it with HMTL via similar settings in Section 4.2. Fig 8 depicts the AFLW2000 error rate on different epochs and the detailed information is reported in Table 5. We observe that puzzled images outperform to normal images and the performance is between SL and HMTL in term of average error rate.

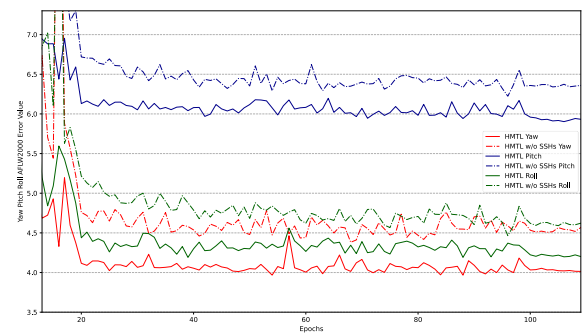

Fig 8 Error rate of HMTL and HMTL w/o SSHs methods on the reduced versions of AFLW2000. HMTL w/o SSHs is similar to the HopeNet method which is only trained puzzled images. Both methods are trained under same settings: learning rate, learning rate decay schedule, regularization, augmentation level.

Table 5 Results of ablation study on the reduced version of AFLW2000. The SL is identical to HopeNet. All the trainings are done with the same settings.

| Method | Yaw (MAE) | Pitch (MAE) | Roll (MAE) | Avg |
|---|---|---|---|---|
| SL | 6.221 | 5.569 | 3.984 | 5.258 |
| HMTL 3×3 puzzling w/o SSHs | 4.589 | 6.223 | 4.465 | 5.092 |
| HMTL 3×3 puzzling | 3.874 | 5.929 | 4.416 | 4.74 |

## 5.2 Hybrid multi-task learning on limited subjects

In this section, the impact of adding self-supervised auxiliary tasks on supervised learning based on the limited number of subjects is going to be examined. For this purpose, ETH-XGaze images with head pose labels have been used. It consists of 80 subjects in the training set. Like Section 4.1, we separate three subjects as the validation set, i.e., 108, 109 and 111. Here, we have tried to show the effect of auxiliary tasks on low data regime compared to SL alone.

For SL procedure, ResNet50 is selected as the backbone. Also, by considering the possibility of the HopeNet losses on the ablation, we have removed it and only used simple regression heads on top of the backbone for yaw and pitch outputs (Equation 3). For the HMTL approach, similar to Section 4.1, the proposed architecture with auxiliary puzzling losses has been selected. Also, beside those experiments, we have fed SL with HMTL inputs (i.e., puzzled images) to evaluate the impact of puzzling perturbation, on the supervised head pose estimation. For all three approaches we initialize the ResNet50 with identical random weights to have a fair comparison.

Random central zoom, random Gaussian noise, cutout, random hue, random brightness and contrast transformation have been used to augment input images heavily. Results are reported in Table 6 and visualized in Fig 9. It is revealed that although feeding puzzled images into supervised HopeNet architecture had beneficial effect, when simple regression heads are used, it shows a negative consequence on the average MAE.

$$\mathcal{L}_{Total} = \mathcal{L}_{SL} + \mathcal{L}_{SSL} = (\mathcal{L}_{Yaw} + \mathcal{L}_{Pitch}) + \mathcal{L}_{puzzle} =$$

$$\left(\left(\sqrt{\frac{1}{n}\sum_{j=1}^{n}(\hat{y}-y)^2}\right)_{yaw} + \left(\sqrt{\frac{1}{n}\sum_{j=1}^{n}(\hat{y}-y)^2}\right)_{pitch}\right)$$

$$+ \sum_{i}^{n}\left(\sum_{part\,j} y_{i,part\,j} \log(\hat{y}_{i,part\,j})\right)_{puzzle}, (3)$$

Where:

- $\mathcal{L}_{SL}$: two RMSE loss for supervised heads
- $\mathcal{L}_{SSL}$: categorical cross entropy for every puzzling self-supervised heads
- $y$: true label
- $\hat{y}$: predicted label
- $n$: number of samples
- $part$: local regions of each image

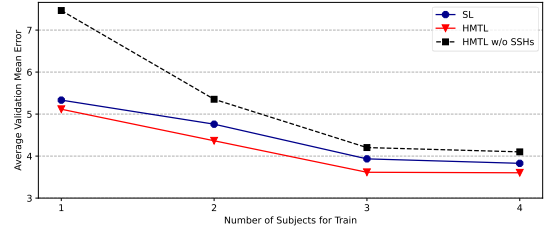

Fig 9 Mean average error of yaw and pitch errors for different number of subjects.

Table 6 Results of mean average errors for Yaw and Pitch via different number of human subjects. Results of this table are the same with Fig 9.

| Method | Subjects | Yaw (MAE) | Pitch (MAE) | Avg |
|---|---|---|---|---|
| SL | 1 | 4.618 | 6.048 | 5.333 |
| SL | 2 | 4.69 | 4.83 | 4.76 |
| SL | 3 | 3.738 | 4.130 | 3.934 |
| SL | 4 | 3.264 | 4.392 | 3.828 |
| HMTL 2×2 puzzling | 1 | 3.739 | 6.493 | 5.116 |
| HMTL 2×2 puzzling | 2 | 3.495 | 5.235 | 4.365 |
| HMTL 2×2 puzzling | 3 | 3.023 | 4.207 | 3.615 |
| HMTL 2×2 puzzling | 4 | 3.12 | 4.086 | **3.603** |
| HMTL 2×2 puzzling w/o SSHs | 1 | 6.6 | 8.332 | 7.466 |
| HMTL 2×2 puzzling w/o SSHs | 2 | 4.746 | 5.96 | 5.353 |
| HMTL 2×2 puzzling w/o SSHs | 3 | 3.490 | 4.916 | 4.203 |
| HMTL 2×2 puzzling w/o SSHs | 4 | 3.705 | 4.498 | 4.101 |

## 6. Discussion

In this article, we tried to answer two questions:

- **How SSL can be utilized for fine-grained head pose estimation?**
- **How architecture design can impact on fine-grained head pose estimation when adding self-supervised auxiliary tasks?**

For the first question, we showed that there are mainly two strategies to use SSL: pre-training it for fine-tuning or using SSL as auxiliary losses besides a SL problem. Both methods were revealed to be able to reduce error rates but the latter showed better results.

From another view, when we use self-supervised auxiliary, the separation points of self-supervised branches are very important. Previous studies [17,19] indicated the effectiveness of MTL. It was shown that some heads could help each other while others may hurt one another. We did not know how to exactly find the correlation of tasks and we could just find it practically. Due to the fact that our computation budget was limited, we could not go for Neural Architecture Search (NAS) which was pretty expensive. Therefore, four points were defined as shared representation locations and after those points, self-supervised and supervised layers were separated. Testing different points approximately showed us the best point of dividing. We did this experiment via two self-supervised pre-text losses, puzzling and rotation. Results indicated that using SSL auxiliary losses was more effective than ImageNet pre-training but only by one of the SSL pre-text tasks (puzzling). Indeed, the other one (rotation) slightly hurt the supervised error rates. Even though the rotation self-supervision task could not be as effective as an auxiliary loss, it was surprisingly helpful to be used as a pre-training technique. Furthermore, an interesting point we found was the effect of puzzled images alone on error reduction via SL. In fact, both puzzled images and adding auxiliary losses could help each other to reduce the average error.

Moreover, it was shown the success of our proposed approach was not limited to large amount of data; HMTL could also be effective in low data regime. Our hybrid approach was only applied during training stage and in the evaluation or test phase, all self-supervised branches could be removed from the main network (backbone + SHs). This means that there is no change in supervised architecture after the training stage and the model would experience no additional floating-point operations per second (FLOPS) in inference time.

Though, we encountered with many questions in this paper which can be considered for future research:

1. **Which architecture is the best for self-supervised auxiliary tasks?**
   Even though, we could show the impact of different branching point on SHs error rate, there is much vagueness about which architecture should be used for each self-supervised tasks and which task hurt supervised tasks instead of helping them (negative transfer). We think, this issue can be explored by NAS for the best architecture design and by gradient modulation [18] for conflicting task gradients.

2. **How to deal with different losses in HMTL optimization?**
   Different tasks in the MTL approach mean different losses. e.g., regression and classification losses are not similar in terms of value, learning speed and performance [18]. When a model is to be trained on more than one task, the various task-specific loss functions must be merged into a single aggregated loss function in which the neural network is trained to minimize it. The simplest method, manual loss weighting, was used in this paper, but there are more promising methods like geometric mean of losses [18] and weighting by uncertainty [50] that have the potential to improve the results and prohibit training collapses for different tasks.

3. **How to find the best SSL auxiliary task for helping a supervised task like head pose estimation?**
   By recent progress in SSL techniques, we think, adding the more recent auxiliary tasks to the SL is feasible and may be more powerful than pretext tasks, specifically, a non-contrastive method which does not need negative pairs while training [14,21,51,52]. However, finding or designing methods like self-supervised pre-text tasks is easier and cheaper to pursue and sometime could lead to magnificent result [53].
   This paper indicated that Barlow Twins based pre-training weights reduced the average error rate in HMTL architecture more than random and ImageNet weights. Furthermore, we guess the best SSL methods, like non-contrastive approaches, represent higher and more abstract feature levels than SL representation due to the generalization of those SSL techniques, as seen in DINO [14]. Indeed, in this training setting, SHs can act as auxiliary tasks for self-supervised representation learning.

# 7. Conclusion

This article tried to demonstrate how self-supervised learning could be used for supervised head pose estimation. The results showed that the use of self-supervised learning in different situations could lead to different results. In other words, it was found that some methods are more suitable for transfer learning and the others could be effective when they are considered as an auxiliary loss during the supervised learning, in which the problem changes to the multi-task learning mode. Because this approach jointly uses the self-supervised and supervised tasks together, in this study it has been called Hybrid Multi-Task Learning or HMTL. Therefore, we designed a customized model based on the original HopeNet architecture by connecting self-supervised learning layers to it. Results showed that the combination of both self-supervised pre-training and HMTL was the best. We also demonstrated that with fine-tuning on a self-supervised based pre-trained weights like Barlow Twins, and adding self-supervised auxiliary losses such as puzzling during fine-tuning, the average error rate was reduced up to 23.1%, which is comparable to the state-of-the-art methods. Although our method, HMTL, showed the lowest error rate when we used both SSL approaches jointly, but it could be quite effective even without doing SSL weight pre-training i.e., training from scratch.

# Acknowledgments

We gratefully acknowledge Hadi Pourmirzaei, Mohammad Pourmirzaei and Elahe Ghazalifar for preparing pictures, resource preparation and helpful discussions.

# References

1. Geronimo D, Lopez AM, Sappa AD, Graf T. Survey of pedestrian detection for advanced driver assistance systems. IEEE Trans Pattern Anal Mach Intell. 2009;32(7):1239–58.

2. Chen Z, Liu Z, Hu H, Bai J, Lian S, Shi F, et al. A realistic face-to-face conversation system based on deep neural networks. In: Proceedings - 2019 International Conference on Computer Vision Workshop, ICCVW 2019. 2019.

3. Liu Z, Hu H, Wang Z, Wang K, Bai J, Lian S. Video synthesis of human upper body with realistic face. In: Adjunct Proceedings of the 2019 IEEE International Symposium on Mixed and Augmented Reality, ISMAR-Adjunct 2019. 2019.

4. Mukherjee SS, Robertson NM. Deep Head Pose: Gaze-Direction Estimation in Multimodal Video. IEEE Trans Multimed. 2015;

5. Cao K, Rong Y, Li C, Tang X, Loy CC. Pose-Robust Face Recognition via Deep Residual Equivariant Mapping. In: Proceedings of the IEEE Computer Society Conference on Computer Vision and Pattern Recognition. 2018.

6. Ruiz N, Chong E, Rehg JM. Fine-grained head pose estimation without keypoints. In: IEEE Computer Society Conference on Computer Vision and Pattern Recognition Workshops. 2018.

7. Valle R, Buenaposada JM, Baumela L. Multi-Task Head

Pose Estimation in-the-Wild. IEEE Trans Pattern Anal Mach Intell. 2021;

8. Liu Z, Chen Z, Bai J, Li S, Lian S. Facial pose estimation by deep learning from label distributions. In: Proceedings - 2019 International Conference on Computer Vision Workshop, ICCVW 2019. 2019.

9. Sheka A, Samun V. Knowledge Distillation from Ensemble of Offsets for Head Pose Estimation. arXiv Prepr arXiv210809183. 2021 Aug;

10. Huang B, Chen R, Xu W, Zhou Q. Improving head pose estimation using two-stage ensembles with top-k regression. Image Vis Comput. 2020;

11. Zhou Y, Gregson J. WHENet: Real-time Fine-Grained Estimation for Wide Range Head Pose. arXiv Prepr arXiv200510353. 2020 May;

12. Yang TY, Chen YT, Lin YY, Chuang YY. Fsa-net: Learning fine-grained structure aggregation for head pose estimation from a single image. In: Proceedings of the IEEE Computer Society Conference on Computer Vision and Pattern Recognition. 2019.

13. Grill JB, Strub F, Altché F, Tallec C, Richemond PH, Buchatskaya E, et al. Bootstrap your own latent a new approach to self-supervised learning. Adv Neural Inf Process Syst. 2020;33:21271–84.

14. Caron M, Touvron H, Misra I, Jégou H, Mairal J, Bojanowski P, et al. Emerging Properties in Self-Supervised Vision Transformers. Proc IEEE/CVF Int Conf Comput Vis (pp 9650-9660). 2021 Apr;

15. Chen X, He K. Exploring simple siamese representation learning. In: Proceedings of the IEEE/CVF Conference on Computer Vision and Pattern Recognition. 2021. p. 15750–8.

16. Asano YM, Rupprecht C, Vedaldi A. A critical analysis of self-supervision, or what we can learn from a single image. arXiv Prepr arXiv190413132. 2019;

17. Standley T, Zamir A, Chen D, Guibas L, Malik J, Savarese S. Which tasks should be learned together in multi-task learning? In: 37th International Conference on Machine Learning, ICML 2020. 2020.

18. Crawshaw M. Multi-task learning with deep neural networks: A survey. arXiv preprint arXiv:2009.09796. 2020.

19. Pourmirzaei M, Montazer GA, Esmaili F. Using Self-Supervised Auxiliary Tasks to Improve Fine-Grained Facial Representation. arXiv Prepr arXiv210506421. 2021 May;

20. Noroozi M, Favaro P. Unsupervised learning of visual representations by solving jigsaw puzzles. In: European conference on computer vision. Springer; 2016. p. 69–84.

21. Zbontar J, Jing L, Misra I, LeCun Y, Deny S. Barlow Twins: Self-Supervised Learning via Redundancy Reduction. arXiv Prepr arXiv210303230. 2021;

22. He K, Zhang X, Ren S, Sun J. Deep residual learning for image recognition. In: Proceedings of the IEEE conference on computer vision and pattern recognition. 2016. p. 770–8.

23. Zhu X, Lei Z, Liu X, Shi H, Li SZ. Face alignment across large poses: A 3d solution. In: Proceedings of the IEEE conference on computer vision and pattern recognition. 2016. p. 146–55.

24. Yin X, Yu X, Sohn K, Liu X, Chandraker M. Towards Large-Pose Face Frontalization in the Wild. In: Proceedings of the IEEE International Conference on Computer Vision. 2017.

25. Fanelli G, Dantone M, Gall J, Fossati A, Van Gool L. Random Forests for Real Time 3D Face Analysis. Int J Comput vision, 101(3), pp437-458. 2013;

26. Zhang X, Park S, Beeler T, Bradley D, Tang S, Hilliges O. ETH-XGaze: A Large Scale Dataset for Gaze Estimation Under Extreme Head Pose and Gaze Variation. In: Lecture Notes in Computer Science (including subseries Lecture Notes in Artificial Intelligence and Lecture Notes in Bioinformatics). 2020.

27. Khan K, Khan RU, Leonardi R, Migliorati P, Benini S. Head pose estimation: A survey of the last ten years. Signal Process Image Commun. 2021;99:116479.

28. Hsu HW, Wu TY, Wan S, Wong WH, Lee CY. Quatnet: Quaternion-based head pose estimation with multiregression loss. IEEE Trans Multimed. 2019;

29. Albelwi S. Survey on Self-Supervised Learning: Auxiliary Pretext Tasks and Contrastive Learning Methods in Imaging. Entropy. 2022;24(4):551.

30. Gidaris S, Singh P, Komodakis N. Unsupervised representation learning by predicting image rotations. arXiv preprint arXiv:1803.07728. 2018.

31. Zhang R, Isola P, Efros AA. Colorful image colorization. In: European conference on computer vision. Springer; 2016. p. 649–66.

32. Chen T, Kornblith S, Swersky K, Norouzi M, Hinton G. Big Self-Supervised Models are Strong Semi-Supervised Learners. Advances in neural information processing systems, 33, pp.22243-22255. 2020.

33. Cole E, Yang X, Wilber K, Mac Aodha O, Belongie S. When does contrastive visual representation learning work? arXiv Prepr arXiv210505837. 2021;

34. Wu D, Li S, Zang Z, Wang K, Shang L, Sun B, et al. Align Yourself: Self-supervised Pre-training for Fine-grained Recognition via Saliency Alignment. arXiv Prepr arXiv210615788. 2021 Jun;

35. Mustikovela SK, Jampani V, De Mello S, Liu S, Iqbal U, Rother C, et al. Self-Supervised Viewpoint Learning from Image Collections. In: Proceedings of the IEEE Computer Society Conference on Computer Vision and Pattern Recognition. 2020.

36. Yun K, Park J, Cho J. Robust Human Pose Estimation for Rotation via Self-Supervised Learning. IEEE Access. 2020;

37. Kundu JN, Seth S, Jampani V, Rakesh M, Venkatesh Babu R, Chakraborty A. Self-supervised 3D human pose estimation via part guided novel image synthesis. In: Proceedings of the IEEE Computer Society Conference on Computer Vision and Pattern Recognition. 2020.

38. Dahiya A, Spurr A, Hilliges O. Exploring self-supervised

learning techniques for hand pose estimation. In: Bertinetto L, Henriques JF, Albanie S, Paganini M, Varol G, editors. NeurIPS 2020 Workshop on Pre-registration in Machine Learning. PMLR; 2021. p. 255–71. (Proceedings of Machine Learning Research; vol. 148).

39. Zhai X, Oliver A, Kolesnikov A, Beyer L. S4L: Self-Supervised Semi-Supervised Learning. Proc IEEE/CVF Int Conf Comput Vis (pp 1476-1485). 2019 May;

40. Korbar B, Tran D, Torresani L. Cooperative Learning of Audio and Video Models from Self-Supervised Synchronization. Adv Neural Inf Process Syst 31. 2018 Jun;

41. Han T, Xie W, Zisserman A. Self-supervised co-training for video representation learning. In: Advances in Neural Information Processing Systems, 33, pp5679-5690. 2020.

42. Oord A van den, Li Y, Vinyals O. Representation learning with contrastive predictive coding. arXiv Prepr arXiv180703748. 2018;

43. Zhuang J, Tang T, Ding Y, Tatikonda S, Dvornek N, Papademetris X, et al. AdaBelief optimizer: Adapting stepsizes by the belief in observed gradients. Advances in neural information processing systems, 33, pp.18795-18806. 2020.

44. Srivastava N, Hinton G, Krizhevsky A, Sutskever I, Salakhutdinov R. Dropout: A simple way to prevent neural networks from overfitting. J Mach Learn Res 15(1), pp1929-1958. 2014;

45. Kazemi V, Sullivan J. One millisecond face alignment with an ensemble of regression trees. In: Proceedings of the IEEE Computer Society Conference on Computer Vision and Pattern Recognition. 2014.

46. Deng J, Guo J, Ververas E, Kotsia I, Zafeiriou S. Retinaface: Single-shot multi-level face localisation in the wild. In: Proceedings of the IEEE Computer Society Conference on Computer Vision and Pattern Recognition. 2020.

47. Wang H, Chen Z, Zhou Y. Hybrid coarse-fine classification for head pose estimation. arXiv Prepr arXiv190106778. 2019;

48. Cao Z, Chu Z, Liu D, Chen Y. A vector-based representation to enhance head pose estimation. In: Proceedings - 2021 IEEE Winter Conference on Applications of Computer Vision, WACV 2021. 2021.

49. Albiero V, Chen X, Yin X, Pang G, Hassner T. img2pose: Face alignment and detection via 6dof, face pose estimation. In: Proceedings of the IEEE/CVF Conference on Computer Vision and Pattern Recognition. 2021. p. 7617–27.

50. Chennupati S, Sistu G, Yogamani S, Rawashdeh SA. MultiNet++: Multi-stream feature aggregation and geometric loss strategy for multi-task learning. In: IEEE Computer Society Conference on Computer Vision and Pattern Recognition Workshops. 2019.

51. He K, Chen X, Xie S, Li Y, Dollár P, Girshick R. Masked autoencoders are scalable vision learners. In: Proceedings of the IEEE/CVF Conference on Computer Vision and Pattern Recognition. 2022. p. 16000–9.

52. Bao H, Dong L, Wei F. BEiT: BERT Pre-Training of Image Transformers. arXiv Prepr arXiv210608254. 2021;

53. Cipolla R, Gal Y, Kendall A. Multi-task Learning Using Uncertainty to Weigh Losses for Scene Geometry and Semantics. In: Proceedings of the IEEE Computer Society Conference on Computer Vision and Pattern Recognition. 2018.